\documentclass[sigconf]{acmart}
\usepackage{graphicx}
\usepackage{soul}
\usepackage{multirow}
\usepackage{multicol}
\usepackage{colortbl}
\usepackage{xcolor,soul,framed}
\usepackage{arydshln}

\usepackage{enumitem}
\usepackage{colortbl}
\usepackage{amssymb}

\usepackage{amsmath}

\usepackage{subfig}
\usepackage{hyperref}

\usepackage{tikz}
\usetikzlibrary{shapes,snakes,positioning,backgrounds,calc,tikzmark, fit}
\usepackage{pgfplots}
\usepgfplotslibrary{groupplots}
\usepackage{caption}
\usepackage{algorithm}
\usepackage{algorithmic}

\newcommand{\name}{\texttt{NACL}}
\copyrightyear{2022}
\acmYear{2022}
\setcopyright{acmcopyright}\acmConference[KDD '22]{Proceedings of the 28th ACM SIGKDD Conference on Knowledge Discovery and Data Mining}{August 14--18, 2022}{Washington, DC, USA}
\acmBooktitle{Proceedings of the 28th ACM SIGKDD Conference on Knowledge Discovery and Data Mining (KDD '22), August 14--18, 2022, Washington, DC, USA}
\acmPrice{15.00}
\acmDOI{10.1145/3534678.3539161}
\acmISBN{978-1-4503-9385-0/22/08}


\begin{document}

\title{Proactively Reducing the Hate Intensity of Online Posts via {\em Hate Speech Normalization}}

\author{Sarah Masud$^*$, Manjot Bedi$^{\ddagger}$, Mohammad Aflah Khan$^*$, Md Shad Akhtar$^*$, Tanmoy Chakraborty$^*$}
 \affiliation{%
   \country{
   $^*$Dept. of CSE, IIIT-Delhi, India; $^{\ddagger}$Northeastern University, USA}
 }
\email{{sarahm,aflah20082,shad.akhtar,tanmoy}@iiitd.ac.in,bedi.m@northeastern.edu}
\renewcommand{\shortauthors}{Sarah Masud, Manjot Bedi, Mohammad Aflah Khan, Md. Shad Akhtar, Tanmoy Chakraborty}

\begin{abstract}
Curbing online hate speech has become the need of the hour; however, a blanket ban on such activities is infeasible due to several geopolitical and cultural reasons. To reduce the severity of the problem, in this paper, we introduce a novel task, {\em hate speech normalization} that aims to weaken the intensity of hatred exhibited by an online post. The intention of hate speech normalization is not to support hate but instead to provide the users with a stepping stone towards non-hate while giving online platforms more time to monitor any improvement in the user's behavior.

To this end, we manually curated a parallel corpus -- hate texts and their normalized counterparts (a {\em normalized text} is less hateful and more benign). We introduce \name, a simple yet efficient hate speech normalization model that operates in three stages -- first, it measures the hate intensity of the original sample; second, it identifies the hate span(s) within it; and finally, it reduces hate intensity by paraphrasing the hate spans.\footnote{{\color{red} \noindent\textbf{Disclaimer:} This paper includes examples of hate speech which contain some profane words. These examples are only included for contextual understanding. We tried our best to censor vulgar, offensive, or hateful words. We assert that we do not support these views in any way.}} We perform extensive experiments to measure the efficacy of \name\ via three-way evaluation (intrinsic, extrinsic, and human-study). We observe that \name\ outperforms six  baselines --
\name\ yields a score of $0.1365$ RMSE for the intensity prediction, $0.622$ F1-score in the span identification, and $82.27$ BLEU and $80.05$ perplexity for the normalized text generation. We further show the generalizability of \name\ across other platforms (Reddit, Facebook, Gab). An interactive prototype of \name\ was put together for the user study. Further, the tool is being deployed in a real-world setting at Wipro AI as a part of their mission to tackle harmful content on online platforms.
\end{abstract}
\maketitle

\section{Introduction}
\label{sec:intro}
Today, anyone with access to the Internet can learn, share, and participate in matters that concern them. However, anti-social behavior such as spreading fake news, cyberbullying, polarization, and posting hateful messages has increased with Internet penetration. While what constitutes hate speech is subjective, there is a common understanding of what underpins hate \cite{doi:10.1177/1754073917751229}. Hate speech can be understood as any communication in speech, writing, or behavior that attacks/discriminates or uses pejorative language against an individual or a group based on who they are. The attack is often targeted at religion, ethnicity, nationality, race, color, descent, gender, or other identity factors \cite{waseem-hovy-2016-hateful}. Our experiment aims to capture varying degrees and designs of hatefulness; we use hate as an umbrella term encompassing overlapping definitions of hate, abuse, and offense. Hate speech on social media usually consists of content that is often in violation of the terms of usage of the platform and leads to either flagging of content or banning of users or both. This has led to an influx of banned users turning over to less-restrictive platforms \cite{bigt}. A cross-platform study of hateful users revealed how the banning of users could backfire for the platform and the Internet community at large. \cite{Johnson2019}. \begin{table*}[ht!]
    \centering
    \caption{Data samples with hate intensity and spans. Intensity scales from $[1-10]$ (a higher score denotes a higher degree hatefulness in the sample). The phrases in blue denote the hate spans in the original sentence. We also provide the respective normalized version of the samples. Section \ref{sec:dataset} describes the annotation scheme and hate intensity score in details (\textbf{Int:} Intensity).}
    \label{tab:dataset:examples}
    \resizebox{\textwidth}{!}
    {
    \begin{tabular}{l|p{52em}|c}
    \hline
    \multicolumn{1}{c|}{\bf Type} & \multicolumn{1}{c|}{\bf Sample} & \bf Int  \\ \hline \hline
    {Org } & {\em This \{\textcolor{blue}{immigrant should be hung or shot ! Period ! An***}\}\textcolor{red}{$_{Span}$} . @user} & 8   \\ \hdashline 
    {Norm} & This immigrant should be punished . @user & 4  \\ \hline

     {Org} & {\em @user This \{\textcolor{blue}{trashy wh*** gold digger}\}\textcolor{red}{$_{Span}$} at again trying be something that makes her think shes good ughhh not happening \{\textcolor{blue}{h** bag pos}\}\textcolor{red}{$_{Span}$}} & 9   \\ \hdashline 
    {Norm} & @user This money loving women at again trying be something that makes her think shes good ughhh not happening & 2  \\ \hline  
    
     {Org} & {\em \{\textcolor{blue}{\#BuildThatWall \#BuildTheDamnWall \#BuildThatWall  BUILD THAT EFFEN WALL}\}\textcolor{red}{$_{Span}$} I'm sorry my Lord \#Jesus but people are just deaf down here} & 8   \\ \hdashline
    {Norm} & Ask them to leave the nation. I'm sorry my Lord Jesus but people are just deaf down here & 4  \\ \hline
    
     {Org} & {\em Women ... Can't live with them ... \{\textcolor{blue}{Can't shoot them}\}\textcolor{red}{$_{Span}$}} & 6   \\ \hdashline 
    {Norm} & Women ... Can't live with them ... can't speak to them & 3  \\ \hline
    
\end{tabular}}
\end{table*}

\textcolor{black}{\textbf{Motivation and evidence.} An alternative solution is to proactively counter \cite{chaudhary2021countering} hate speech before it goes public. \emph{Systematic and proactive sensitization of online users can help them voice their opinions without directly propagating harm}. In this case, we believe that promoting users with alternative scenarios nudges a change in attitude \cite{PMID:22448888, 10.1145/3272973.3274056}. Twitter and Instagram, among other platforms, have already rolled out the experimental prompts for `offensive posts with an option to revise before going public' \cite{Twitter,Instagram}. These tools state that the material is objectionable and leave it to the users to build a more appropriate form. Based on experiments with such prompts, researchers at Twitter published findings \cite{twitter_study} that prompted participants to eventually post fewer offensive tweets than non-prompted users in the control group. Such nudging should be more welcoming to the users than them getting banned from the platform without having an opportunity to improve. 
}

\textbf{Contribution I: Novel problem statement.} Prompting users to change their texts to be completely non-hateful can be a significant behavioral shift and may backfire. Therefore, as an experiment, we take the first step of making the content less hateful and hope to build from there. Extending upon the idea of proactive prompts, we propose the novel task of \emph{hate speech normalization}, which \textit{suggests} a normalized counterpart of hateful content by weakening the overall hatred exhibited by the original content while retaining the underlying semantics. Additionally, in line with previous studies \cite{twitter_study}, we observe a significant reduction in user engagement (virality) for normalized sentences compared to the original one. Our hypothesis testing is detailed in Section \ref{sec:hypo}.

\textbf{Contribution II: New dataset.} Note that our aim is {\em not} to render a hateful message into a non-hateful one; instead, it is the \textit{reduction of hatred}. To this end, we manually curate a parallel corpus of hateful samples and their normalized versions, {\em the first dataset of its kind}. Our preliminary analysis of hateful samples suggests that, in general, only a few phrases within a sentence convey major hatred. Therefore,  we first identify the hateful spans for each hateful sample and then normalize them to reduce overall hatred. Table \ref{tab:dataset:examples} lists examples of such posts and their normalized counterparts.

\textbf{Contribution III: Novel method.} We then propose \textbf{\name}, a {\bf N}eural h{\bf A}te spee{\bf C}h norma{\bf L}izer, which operates in three stages. It first measures the intensity of hatred exhibited by an incoming sample by employing an attention-driven BiLSTM-based regression model. Following this, a BiLSTM-CRF module operates on the sample to extract the hateful spans from the source sample. In the last stage, we incorporate and fine-tune a BART-based sequence-to-sequence model (seq2seq) \cite{lewis-etal-2020:BART} to normalize the identified hate spans with feedback (reward/penalty) from the hate intensity predictor-driven discriminator module. The reward/penalty from the discriminator enforces the generator (BART model) to paraphrase the hate spans to reduce the hate intensity. For each stage, we compare our results with various baselines. We develop an end-to-end system that can be efficiently deployed in the real-world environment. It achieves consistent performance in offline (intrinsic/extrinsic evaluations) as well as online evaluations (as observed by annotators).  

\textbf{Contribution IV: Extensive evaluation and deployment.} We perform an extrinsic evaluation on three standard state-of-the-art hate detection models where we measure the confidence of these models in classifying the normalized samples as `\textit{hate}' against the original unnormalized samples. We hypothesize that the models should have lower `\textit{hate}' confidence for the normalized (reduced hate) sample compared to the original one. We further compare \name\ with six baselines and obtain a performance of $0.136$ RMSE, $0.622$ F1-score, and $82.27$ BLEU in intensity prediction, span identification, and text normalization, respectively.

In partnership with Wipro AI, we develop an interactive web interface and conduct an anonymized human evaluation on the outputs of \name\ and other baselines. We observe that \name\ reports better results in intrinsic evaluation, higher reduction in confidence of hate detection models in extrinsic evaluation, and coherent and better outputs as evaluated by humans. We further show the {\em generalizability} of \name\ across three social media platforms -- Reddit, Facebook, and Gab.

\noindent\textbf{Reproducibility:} Being aware of the ethical considerations (Appendix \ref{app:ethics}) of dataset, we are restricting its release. Meanwhile, the models, source codes, annotation guidelines, the questionnaires used for human evaluation (c.f. Section \ref{sec:human}), and web-tool (c.f. Section \ref{sec:browser}) is available at \url{https://github.com/LCS2-IIITD/Hate_Norm}.

\section{Application and Target Audience}
\label{sec:hypo}
\textcolor{black}{\textbf{Target Audience.} People initiate and engage in hateful conversation for a variety of reasons. There will always be users who intentionally and constantly spread hateful content. The majority of such users eventually get flagged and reported by the community for their anti-social behavior. However, despite intentionally engaging in hateful content, a section of online users are adaptive and can be nudged to change their opinions via empathy \cite{Hangartnere2116310118} and corrective behavior. Our tool is aimed at the latter set of user groups whose social engagements can be eventually guided to become non-hateful. Meanwhile, prompting tools and the adoption of suggestive texts can help the content moderators understand the linguistic behavior of offensive users and track any changes in user behavior against getting flagged/banned.  
}

\begin{figure}[!t]
\centering
    \includegraphics[width=\columnwidth]{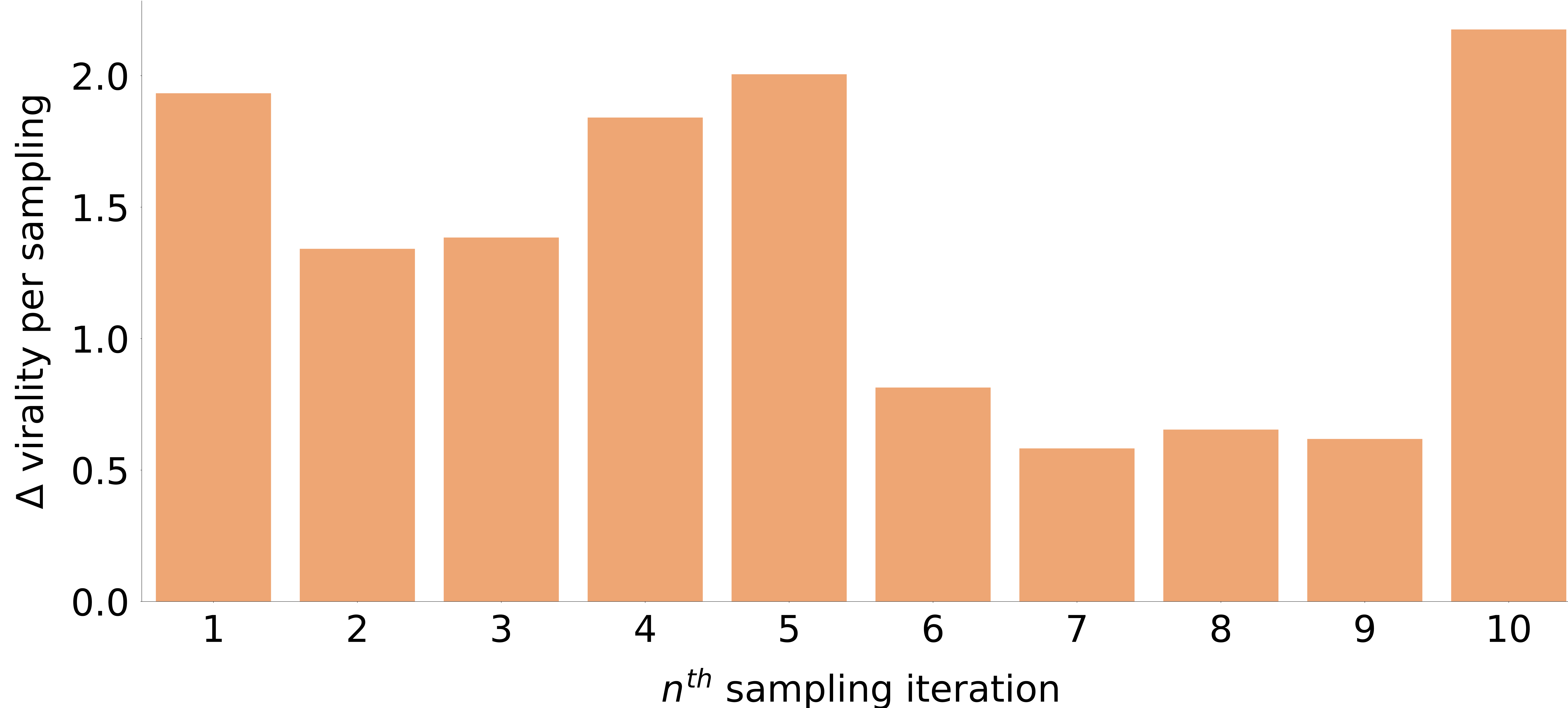}
\caption{Difference in predicted number of comments per set per iteration. During each iteration, we obtain the median difference ($\Delta$) in the predicted number of comments for hateful and corresponding normalized sampled pairs.}
\label{fig:hypo}
\vspace{-5mm}
\end{figure}

\textcolor{black}{\textbf{Evidence.}
A significant criticism for prompting and suggestive rectification is the debate of reduced expressive powers of the users. However, a recent study by Twitter \cite{twitter_study} also revealed that encouraging users to reduce the publishing of offensive text had no significant impact on their ability to participate in the non-offensive conversation. This result is a big motivation for researchers exploring proactive methods of countering hateful/hurtful content, as we do in this paper.
Though one can argue that normalized text is still hateful and prone to spreading harm, we hypothesize that \textit{in its normalized form with a reduction in intensity, the content should see a decline in the user engagement it receives} (i.e., reduces its virality). We take inspiration from the virality prediction models on social media to test this hypothesis. In our setting, the virality of a post is expressed in terms of the total comments it receives. Since ours is a text-only dataset curated from different sources, we lack the availability of any network, temporal and other signals. Engineering various textual features (listed in Appendix \ref{app:viral}) inspired by the work of \citet{10.1145/3394486.3403251} and \citet{10.1145/2566486.2567997}, we train a user engagement prediction model (trained on Reddit dataset). We use it to predict the comment count for the hateful posts and their normalized counterpart present in our dataset. We use hateful and normalized samples as our control and alternate sets. We randomly sample $300$ pairs of test cases for $10$ iterations and record differences in median comment count per sampling iteration (depicted in Figure \ref{fig:hypo}). We found the phenomenon ($\Delta$ in the median virality of hate vs. normalized) to be statistically significant with a $p$-value of $0.0027$ and effect-size of $2.324$ on Welch's t-test. Thus, we conclude that even if present online, the normalized form of a post is less likely to gain engagement and go viral. This evidence is in line with existing studies \cite{twitter_study} which found that less offensive content is expected to catch fewer eyeballs.
}

\section{Objective \& Individual Stages}
\label{sec:formal_def}
For a given hate sample $t$, our objective is to obtain its normalized form $t^\prime$ such that the intensity of hatred $\phi_t$ is reduced i.e., $\phi_{t^\prime} < \phi_t$. To achieve this, we divide the overall task into three stages.
\begin{itemize}[noitemsep, leftmargin=*]
    \item \textbf{Hate Intensity Prediction ({{HIP}).}} The task of hate intensity prediction is to determine the extent (degree or intensity) of hatred in a message (inspired by the idea of measuring online toxicity\footnote{https://www.perspectiveapi.com}). Given a sample $t$, it measures the intensity of hatred on a scale of $[1, 10]$, with $10$ being the highest, i.e., $\phi_t = HIP(t)$. 
    \item \textbf{Hateful Span Identification ({HSI}).} Hateful spans in a sentence are the portions of a sentence responsible for conveying hate \cite{pavlopoulos-etal-2021-semeval}. The intuition behind this task is that we can achieve sentence-level normalization if we attempt to normalize these spans. 
    \emph{A sample can have multiple, non-overlapping hate spans, and $HSI(t)$ aims to find all such spans.}
    \item \textbf{Hate Intensity Reduction ({HIR}).} The objective of  hate intensity reduction is to generate a new sample that preserves the original semantic but with weaker hate intensity. Intuitively, we first need to identify the threshold for strong vs weak hate. For our experiments, we set a threshold $\tau = 5$. For each strong hate sample with $\phi_t > \tau$, we aim to generate semantically similar sample $t^\prime \approx t$, with the constraint $\phi_{t^\prime} \le \tau$.
\end{itemize}

\begin{figure}[t!]
    \centering
    \includegraphics[width=\columnwidth]{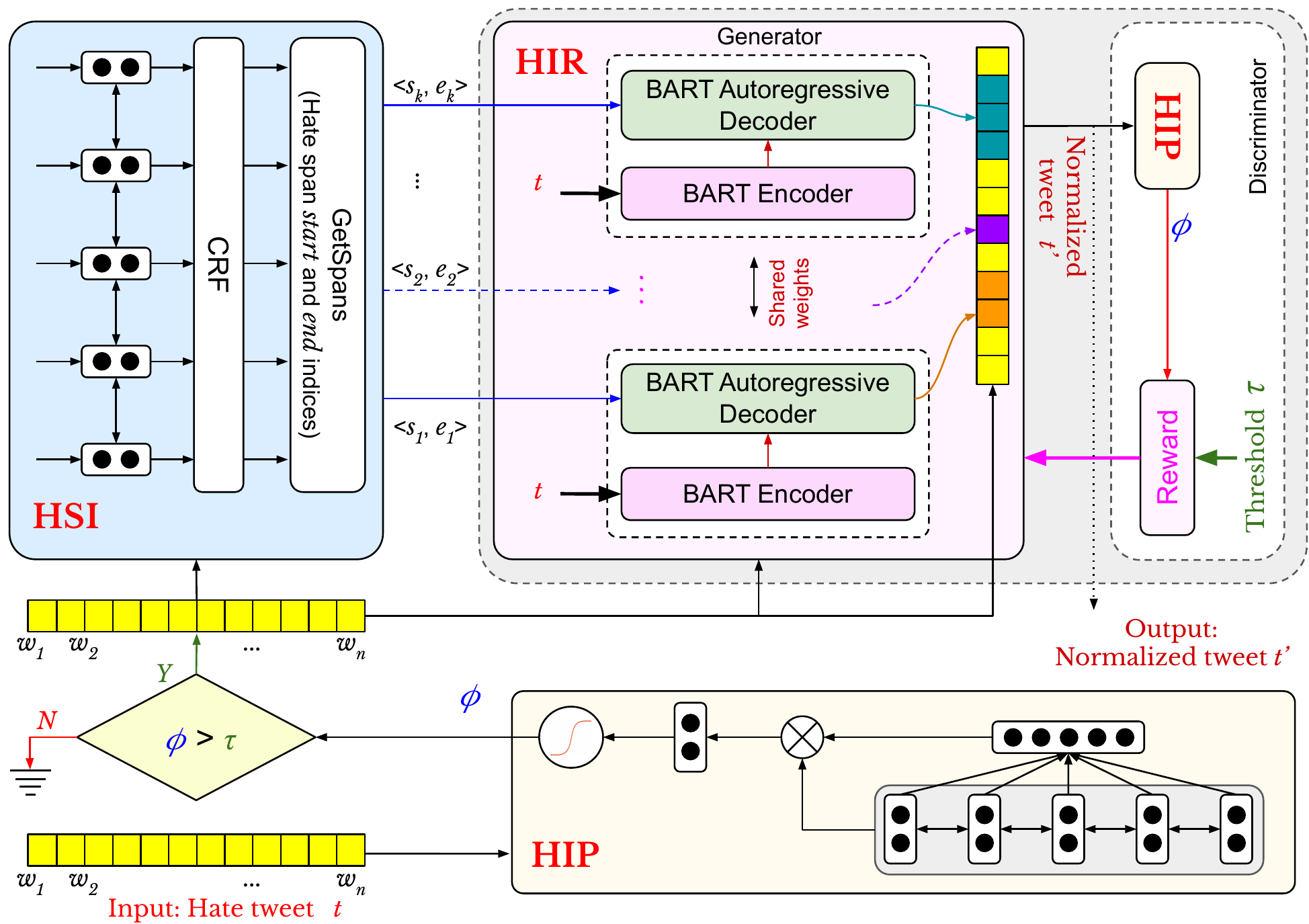}
    \caption{A schematic view of \name. The first stage of the model (i.e., HIP) validates the eligibility of a sample for normalization. Subsequently, HSI identifies the start and the end indices to mark hate spans in eligible samples. Finally, for each identified hate span, a normalized paraphrase is generated by HIR and validated by the discriminator to ensure the intensity reduction. 
}
    \label{fig:architecture}
    \vspace{-13pt}
\end{figure}

\section{Proposed Methodology}
We engineer a real-time system that can normalize hate speech on the fly. Our goal is to design a simple yet effective industry-level solution that can be deployed at scale. Interestingly from our experimentation, we observe that a simpler architecture produces comparable performance to a complex solution and is more industry-centric in terms of response time and performance.  

Figure \ref{fig:architecture} illustrates a high-level overview of the \name\ framework. We assume that \name\ will be called to action once a text is detected as hateful by a preexisting hate detection model. In our case, we mimic it by only using hateful samples for training \name\ (c.f. Section \ref{sec:dataset}). Any future reference of the input sample would mean that it is already labeled as hate. In the first step, we compute the hate intensity ($\phi_t$) of an input sample $t$ and forward it for normalization only if it satisfies the condition, $\phi_t > \tau$ (i.e., only strongly hateful samples are subject to normalization). Next, the model predicts the hate spans within the incoming sample following the CONLL-2002 \textit{BIO} notations \cite{tjong-kim-sang-2002-introduction}. Subsequently, \name\ normalizes each identified hate span through a generator module. Finally, the outputs of each generator are interleaved in the original span sequence and fed to the discriminator to ensure the reduction of hate intensity ($\phi \leq \tau$). The pseudo-code is summarized in Appendix \ref{app:algo}. 

\textbf{\name-{HIP} Module:}
For hate intensity prediction, we employ a Bi-LSTM model. The intermediate representation is then passed through a self-attention layer connected to a fully-connected layer with linear activation to obtain the input sample's continuous hate intensity prediction. Table \ref{tab:results:strength} shows various models and pre-trained embeddings we experimented with.

\textbf{\name-{HSI} Module:}
For the span detection model, we initially take as input a sequence of a hateful sample ($X_i$). For every token of the sequences, we have manually annotated span labeling that represents the start and end of a hateful span in it. Our goal is to train our model to detect relevant hate spans represented by tags $(B, I, O)$ where 'B' represents the beginning of hate span, 'I' forms the continuation of a hate span, and 'O' represents the non-hate tag. For our model, we use bidirectional LSTMs to capture the contextual representation of sequences. The hidden representations are then passed through a Time Distributed dense layer to flatten the embedding structure. We further use a CRF model layer to fit our representations and produce the required tags for our sequences. 
The CRF layer models the conditional probability of each sequence $p(s_1, ...., s_m|x_1, ...., x_m)$ by defining a feature map $\phi(x_1,...,x_m,s_1,...s_m) \in R^d $ that maps an entire sequence to a d dimensional feature vector. Then we can model the probability using the feature vector $w \in R^d$ as follows:
\begin{eqnarray}
 p(s|x;w)=\sum_s exp(w.\phi(x,s)),
\end{eqnarray}
where s ranges across different sequences and x represents the POS tag. After getting the feature vector $w^*$, we can find the most likely tag for a sentence $s^*$ by:
$s^*=argmax$ $p(s|x;w*)$

\textbf{\name-{HIR} Module:}
The hate intensity reduction model is a Generative Adversarial Network (GAN) based architecture, which employs both \name-{HIP} and \name-{HSI} as its assistive modules. \name-{HIR}  accepts a hateful sample along with its span labels as identified by the {HSI} model. For \name-{HIR}, we use a pre-trained BART \cite{lewis-etal-2020:BART} and fine-tune it for the hate normalization task. The BART-based seq-to-seq model generates the normalized spans based on the hate spans identified by the \name-{HSI} module. The generated spans are amalgamated with the rest of the sample tokens and forwarded to the {HIP}-based discriminator model.

The discriminator computes the hate intensity score $\phi_{t^\prime}$ of the normalized text $t^\prime$. Since our objective is to reduce the intensity of $t^\prime$, we assign a reward/penalty $R$ to the generator based on the hate intensity score $\phi_{t^\prime}$ of the normalized sentence and the accepted threshold value $\tau$ as follows:
\begin{eqnarray}
R_{t^\prime} = \tau - \phi_{t^\prime}
\end{eqnarray}
Note that the threshold value is a hyperparameter. In this paper, we experiment with two threshold values ($\tau=\{3,5\}$) for the hate normalization. Table \ref{tab:results:strength:reduction} shows that $\tau=5$ is better suited for the \name-HIR model. Therefore, for the rest of the experimentation, we continued with $\tau=5$ as the threshold value for our model. If the computed hate intensity score is lesser than or equal to the threshold, the positive reward encourages the generator to continue predicting the similar normalized span;  otherwise, the negative reward penalizes the generator to improve its prediction. The generator consumes the reward into its loss function as follows:
\begin{eqnarray}
\mathcal{L} = \ell + (1 - R)
\end{eqnarray}
where $\ell$ is the generator loss. Backpropagation aims to minimize the consolidated loss $\mathcal{L}$ in order to generate semantically coherent normalized samples with hate intensity lesser than equal to $\tau$.

\textbf{Some Engineering Observations.} As stated at the beginning of this section, we aim to find a solution that not only performs well for the given problem statement but can also be deployed. Subsequently, we make a trade-off between efficiency and complexity at each modeling stage. For instance, adding additional BiLSTM layers for HIP did not significantly boost the performance; hence, we exclude them from the final modeling. Additionally, with HSI, simpler embedding solutions like Glove and ElMo performed comparably to BERT. We think the main reason for BERT's low performance is that the BERT tokenizer splits the words into subtokens, which causes the spans to be distributed unequally across the subtokenized words. This skews the target values that belong to one of the tags ($o_i \in (B, I, O)$), resulting in the model overfitting. As we have limited data, the ELMO model performs better with this data and performs well for the problem. For both HSI and HIR, we make the obvious decision to use the optimized, distilled versions of the larger BERT and BART models (as supported by HuggingFace\footnote{\url{https://huggingface.co/}}), further reducing the number of parameters to train.

\begin{table}[!t]
    \centering
    \caption{Dataset statistics.}
    \label{tab:dataset:stat}
    \scalebox{0.85}{
    \begin{tabular}{l|c}
    \hline
        \textbf{Statistics} & {\bf Value} \\ \hline \hline 
        Total samples & $4423$ \\
        Sample length & $23$ (avg), $112$ (max) \\ \hline 
        No. of samples with intensity scores & $4423$ \\
        Hate intensity range & 1 (min) -- 10 (max)\\ \hline
        No. of samples with spans & $3027$ \\
        No. of spans & $5732$ \\
        Avg length of spans (tokens) & $3$ \\ \hline 
        No. of normalized samples & $3027$ \\
        Normalized sample length & $21$ (avg), $99$ (max) \\ \hline
    \end{tabular}
    }
    \vspace{-5mm}
\end{table}
\section{Dataset}
\label{sec:dataset}
We sourced and compiled a list of hateful instances from \cite{basile-etal-2019-semeval,davidson2019racial,de-gibert-etal-2018-hate,chung-etal-2019-conan,Mathew_Saha_Yimam_Biemann_Goyal_Mukherjee_2021,jha-mamidi-2017-compliment} by restricting our data creation to the hateful labels of the respective datasets. In total, we collected $4423$ hateful samples and annotated them with hate intensity scores and hateful spans. At the end of the process, we observed $1396$ samples as either implicit (no apparent hate span) or exhibiting hate intensity less than the threshold $\tau$. Thus, our final gold dataset consists of $3027$ parallel hateful and normalized samples, along with the intensity scores and hate spans in original ones (Table \ref{tab:dataset:stat}). Additionally, we annotated each normalized sample with its new hate intensity score. Two annotators were employed for this task.

\begin{figure}[!h]
\centering
    \includegraphics[width=\columnwidth]{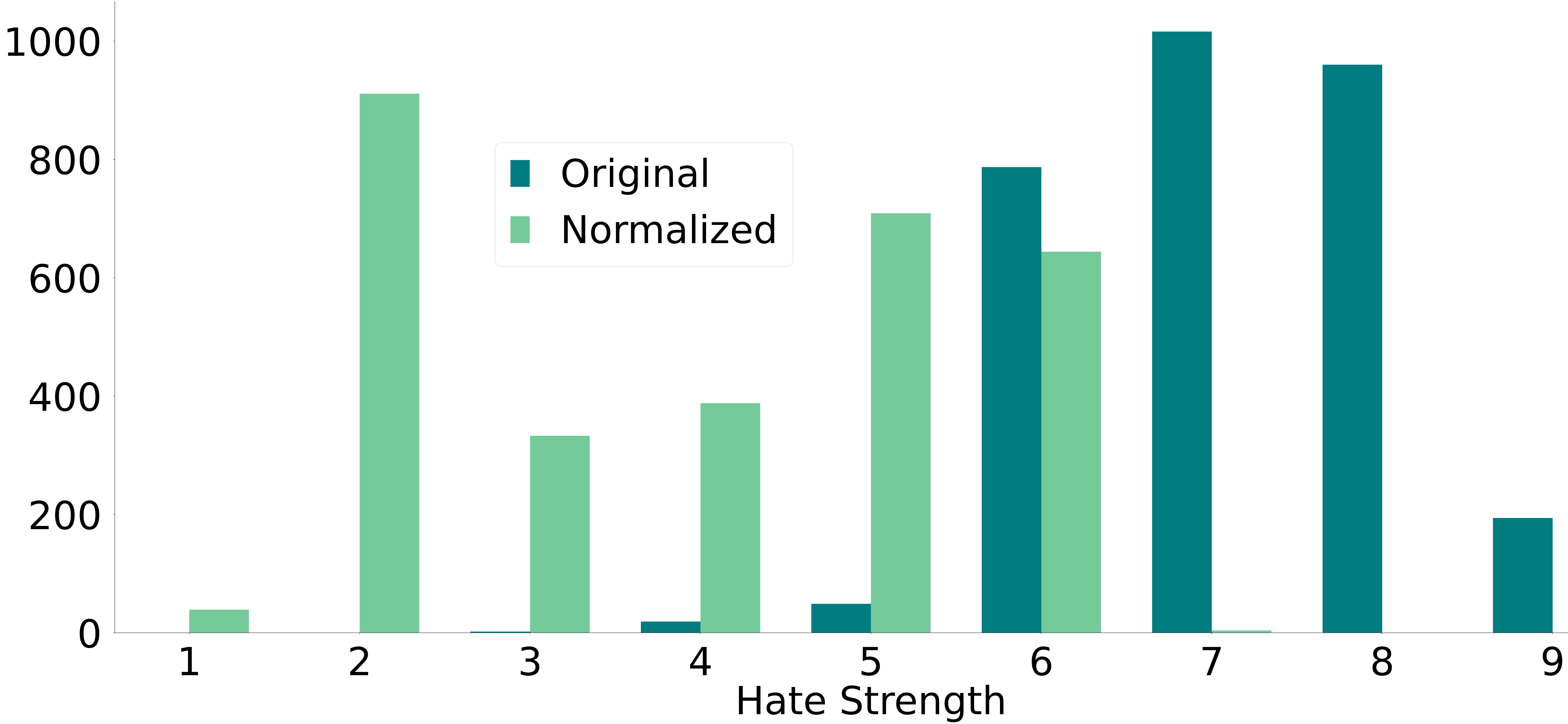}
\caption{Hate intensity distribution for the original and normalized samples. The distribution for original samples shifts towards the weaker intensities in normalized samples.}
\label{fig:histogram}
\vspace{-5mm}
\end{figure}

Both annotators annotated all the available samples. In case of disagreement ($11\%$ of cases), we employed a third annotator to break the tie. Overall the inter-annotator mean squared error for hate intensity was $0.20$. In case of a disagreement on hate-span, if the non-overlap contains words like abuse or racial slurs, we added them to the span. In the rare case, if both annotators and third annotators were in disagreement over a span, the sample was dropped from the final dataset. Table \ref{tab:dataset:stat} lists the overall statistics of the data curated and annotated. The detailed annotation guidelines are covered in Appendix \ref{app:annotation_guideline}. Table \ref{tab:dataset:examples} shows a few examples of the original hate samples, along with their normalized forms and intensity scores for both sets.\footnote{For annotations and human evaluation, the texts were not masked and presented as-is.} A bar graph representing the original and normalized hate intensity distributions is shown in Figure \ref{fig:histogram}. It is evident that the intensity distribution mass in original samples shifts towards the weaker intensity in normalized samples.    

\noindent\textbf{Train and test sets:} 
We split the annotated dataset in the $70:15:15$ for the training ($2119$), validation ($454$), and testing ($455$). One may rightfully argue that such a small number of samples in the test set ($S_A$) may not be adequate to evaluate the modules. Therefore, we compiled another set of $1,111$ hate samples from two sources, released by \citet{ziems2020racism} and \citet{DBLP:journals/corr/abs-1710-07395} as the secondary test set ($S_B$). Note that samples in $S_B$ do not have gold normalized counterparts; therefore, the normalized samples generated from $S_B$ cannot be used for intrinsic evaluations.  

\section{Evaluation and Analyses}
We design a three-way evaluation framework. First, we perform an intrinsic evaluation to measure the model's performance by traditional evaluation metrics. Second, we perform human evaluation where we ask human annotators to rate the quality of the generated output (from our model and the generation baselines). Third, we employ state-of-the-art hate speech detection models to discriminate between the original and normalized hate samples for extrinsic evaluation. Additionally, we evaluate \name\ on datasets from three different platforms -- Reddit, GAB, and Facebook. We further let human emulators supply their content in English and check in real-time the expected output for randomly supplied content. Both these external evaluations help test the robustness of the model against a host of inputs possible in the real world. This is also the first step toward extending the evaluation for adversarial attacks, which we hope to study in the future.
\subsection{\name-{HIR} Baseline Methods}
\label{sec:baseline}
It should be noted that we lack accurate baselines due to the novelty of the proposed task. Therefore, we adopt six existing methods as baselines that we deem pertinent to our work.
\begin{itemize}[noitemsep, leftmargin=1em]
    \item \textbf{Dictionary-based Normalization} \cite{su-etal-2017-rephrasing}: Inspired by the early work of \citet{su-etal-2017-rephrasing}, we define a dictionary-based model that learns the mapping between the hateful span and its corresponding normalized span from the training set. We employ tf-idf based cosine similarity to perform a lookup to select the closest normalized span for the hate span in the test sample.
    \item \textbf{Neutralizing Subjective Bias} \cite{neutralizing:news:bias:2020}: This style-transfer model aims to neutralize the subjective bias in news content by converting a piece of opinionated news into a neutral one. It utilizes an LSTM-based sequence-to-sequence model. We re-train the model for the hate normalization task, assuming the strongly hateful content as biased.
    \item \textbf{FGST} \cite{fgsa:2019:acl}: 
    Given a sentence with a polarity label (\textit{positive} or \textit{negative}), FGST generates a new sentence having a pre-defined sentiment intensity score. In our setting, we aim to generate a normalized text having $\phi_t \leq \tau$.
    \item \textbf{Style Transformer} \cite{dai-etal-2019-style}: 
    Style Transformer is a transformer based architecture which generates a new sentence (for a given source sentence) without making any assumption about the latent representation of source sentence. We re-train the model for style transfer between original and normalized sentences. 
    \item \textbf{Style Transfer (NPTCA)} \cite{NIPS2017_2d2c8394}: 
    Style Transfer (NPTCA) generates a new sentence for a chosen sentence by assuming that different corpora possess a shared latent content distribution. It uses two constrained versions of auto-encoders (aligned and cross-aligned) to refine the alignment of these latent space distributions. We re-train the model for style transfer between original and normalized sentences.
    \item \textbf{Style Transfer (DRG)} \cite{li-etal-2018-delete}: 
    For a sentence, Style Transfer (DRG) generates a new sentence by modifying a specific attribute such as sentiment while keeping other attributes intact. It deletes phrases associated with the attribute in the original text and then retrieves new phrases associated with the target attribute and uses a sequence to sequence model to combine them. For our evaluations, we re-train the model for style transfer between original and normalized sentences.
\end{itemize}
\begin{table}[!t]
\vspace{-7.5mm}
    \centering
    \caption{Results: Hate intensity prediction (HIP).
    }
    \label{tab:results:strength}

    \resizebox{\columnwidth}{!}
    {
    \begin{tabular}{l|c|c|c|c}
    \hline
           \multirow{2}{*}{\bf Model} & \multirow{2}{*}{\bf Description} & \multicolumn{3}{c}{\bf Evaluation Measure}   \\ \cline{3-5}
         & & {\bf Pearson $\uparrow$}   &{\bf  Cosine Sim $\uparrow$} &{\bf  RMSE $\downarrow$} \\ \hline \hline
         CNN & \multirow{3}{*}{GLV} & 0.2827 & 0.2613 & 3.175 \\
         BiLSTM &  & 0.1939  & 0.2044 & 3.411\\
         BiLSTM+CNN &  & 0.3600  & 0.3124 & 4.92\\ \hdashline
         CNN & \multirow{3}{*}{BERT} & 0.2449 & 0.212 & 2.9823 \\
         BiLSTM &  & 0.2613  & 0.2457 & 4.9414\\
         BiLSTM+CNN &  & 0.3211  & 0.3375 &  3.691\\ \hdashline
         BERT & Fine-tune & 0.5558 & 0.5927 & 1.7712\\ \hdashline
         \multirow{3}{*}{\name-{HIP}} & ELMO & 0.4521 & 0.4141 & 0.9982\\
          & BERT (linear) &\textbf{0.766} & \textbf{0.973} & \textbf{0.136}\\
          & BERT (sigmoid) & 0.704 & 0.968 & 0.148\\ \hline
    \end{tabular}}
    \vspace{-5mm}
\end{table}
\vspace{-4mm}
\subsection{Intrinsic Evaluation}
We compute Pearson correlation, cosine similarity and RMSE scores for {HIP} (Table \ref{tab:results:strength}); whereas, we adopt F1-score for HSI (Table \ref{tab:results:span}). In the case of  HIR, which is a  generative task, we compute perplexity and BLEU scores (Table \ref{tab:results:strength:reduction}). The model hyperparameters are covered in Appendix \ref{app:hyper}. We iteratively build our modules, starting independently with the HIP and HSI modules and then using only the best-performing ones for HIR. 

For HIP, as shown in Table \ref{tab:results:strength}, BERT+BiLSTM models performs better than others. Using BERT+BiLSTM with a linear activation gives the best results with cosine similarity of $0.973$, Pearson score of $0.766$ and an RMSE of $0.136$. Meanwhile, BERT+LSTM with a sigmoid activation (scaling intensity between 0-1 range) reports a cosine similarity of $0.968$, Pearson score of $0.704$ and RMSE of $0.148$. As overall the linear activation beats sigmoid, for the rest of our experimentations, we use BERT+LSTM (linear) as our HIP model. 

For  HSI, BiLSTM+CRF yields the best F1 of $0.622$ and best Recall of $0.634$ with ELMo embeddings as reported in Table \ref{tab:results:span}. Among others, the fine-tuned SpanBERT model stands the second-best with $0.583$ F1 and highest precision with $0.6913$. For the rest of our experimentation, we use ELMo+BiLSTM+CRF as our HSI model.

Finally, we report the performance of \name-{HIR} module along with other baselines in Table \ref{tab:results:strength:reduction}. {HIR} yields the best perplexity score (a lower value is better) of $80.05$ for the generated normalized sentences. In comparison, the dictionary-based baseline obtains the perplexity of $92$. For reference, we also compute perplexity ($64.66$) for the reference (gold) normalized sentences. Moreover, we observe a similar trend in the BLEU scores as well. The \name-{HIR} model achieves the highest BLEU ($82.27$). The high BLEU score for {HIR} can be attributed to the fact that we mainly target hate spans for normalization, and a good portion of the original token (not containing hate) sequence gets preserved in the normalized sentence. Categorizing supervised and unsupervised based on whether the method requires parallel data, we observe that supervised methods like Bias Neutralization and some unsupervised methods like FGST perform moderately well in comparison to \name-{HIR}. We observe abysmal performance from most unsupervised style transfer methods. This is mainly due to the lack of a large-scale corpus for training these models for our problem definition. Owing to which most of the generated sequences produce junk values. During the human evaluation, we do not consider these unsupervised baselines except for FGST.
\begin{table}
    \centering
    \caption{Results: Hate span identification (HSI).}
    \label{tab:results:span}
    {
    \resizebox{\columnwidth}{!}
    {
    \begin{tabular}{l|c|c|c|c}
    \hline
           \multirow{2}{*}{\bf Model} & \multirow{2}{*}{\bf Description} & \multicolumn{3}{c}{\bf Evaluation Measure} \\ \cline{3-5}
         & & {\bf Precision $\uparrow$} & {\bf Recall $\uparrow$} & {\bf F1 Score $\uparrow$}   \\ \hline \hline
         CRF & GLV & 0.7013  & 0.3867 & 0.4985 \\ \hdashline
         CRF & BERT & 0.6624  & 0.3335 & 0.4437 \\ \hdashline
         BERT & \multirow{2}{*}{Fine-tune} & 0.6053 & 0.3676 & 0.4574 \\
         SpanBERT &  & 0.7081 & 0.5413 & 0.6135 \\ \hdashline
         \multirow{4}{*}{\name-{HSI}} & GLV & 0.491 & 0.458 & 0.470 \\ 
          & BERT & 0.6471 & 0.4823 & 0.5526 \\ 
          & ELMO & 0.619 & {\bf 0.634} & \textbf{0.622} \\
          & SpanBERT & {\bf 0.6913} & 0.5041 & 0.5830 \\
         \hline
    \end{tabular}}}
    \vspace{-5mm}
\end{table}

\begin{table}
    \centering
    \caption{Results: Hate intensity reduction ({HIR}). 
    }
    \label{tab:results:strength:reduction}
    \resizebox{0.8\columnwidth}{!}
    {
    \begin{tabular}{c|l|c|c}
    \hline
    \multirow{2}{*}{\bf Supervised} & \multirow{2}{*}{\bf Model} & \multicolumn{2}{c}{\bf Evaluation Measure}   \\ \cline{3-4} 
       & & {\bf BLEU $\uparrow$} & {\bf Perplexity $\downarrow$}\\ \hline \hline
       \multirow{2}{*}{Yes} & Dictionary Model & 55.18 & 92\\
         
        & Bias Neutralization & 39.48 &90.38\\ \hdashline
        \multirow{4}{*}{No} & FGST & 39.35 & 123.38\\
        & Style Transformer (ST) & 15.55 & 200.85\\
        & Style Transfer (NPTCA) & 0.93 & 1138.4\\ 
        & Style Transfer (DRG) & 0.84 &199.58\\ \hdashline
       \multirow{2}{*}{Yes} & \name-{HSR ($\tau$=3)} & 58.84  & 86.11\\
        & \name-{HSR ($\tau$=5)} & \textbf{82.27} & \textbf{80.05} \\ \hline
        & Gold & 100 & 64.66\\ \hline
    \end{tabular}}
    \vspace{-5mm}
\end{table}
\subsection{Extrinsic Evaluation}
Our hypothesis for extrinsic evaluation is that if a hate normalization model produces high quality normalized text with lower hate intensity score, then a hate speech detection method will exhibit less confidence in classifying the normalized text as hate. For this evaluation, we employ three widely-used hate speech detection methods -- \cite{waseem-hovy-2016-hateful}, \cite{davidson2019racial}, \cite{founta2018unified}.  We train the methods on their respective datasets. For consistency, we map multiple granular hate labels into \textit{hate} and \textit{non-hate} labels. The original dataset provided by respective papers, is summarized in Appendix \ref{app:ext_eval}.

\noindent\textbf{Evaluation on Test Sets $\mathbf{S_A}$ and $\mathbf{S_B}$:} 
For each original sample $t$ in $S_{A}+S_{B}$, we extract $\gamma(t, m)$, the softmax probability of the \textit{hate} class as the confidence score, where $m$ 
is the underlying hate detection method.
Evidently, $\gamma(\cdot) \in (0,1]$. Subsequently, we compute the confidence score, $\gamma(t^\prime, m)$ for each generated normalized sample $t^\prime$. Considering that hate speech normalization aims to reduce the hate intensity of samples instead of converting to non-hate, we restrict ourselves in analyzing a set of samples in $M_m \subseteq S_{A}+S_{B}$ for which $\gamma(t)\geq0.5 \text{ and } \gamma(t^\prime)\geq 0.5$, i.e., both the original and normalized samples are predicted as hate. Finally, we compute the average difference in confidence score, $\Delta_c$, for each pair $(t,t^\prime)$ in $M_m$ as, 
\begingroup
\setlength\abovedisplayskip{-2pt}
\begin{equation}
    \Delta_c(t,t^\prime) = \frac{1}{|M_m|}\sum_{t \in M_m}\gamma(t) - \gamma(t^\prime)
    \label{eqn:confidence}
\end{equation}
\endgroup
\begin{table}[t!]
\caption{Average change ($\Delta_c$) in confidence of the predicted hate class of three hate detection methods. $\Delta_c$ is computed between the original and normalized pairs in the test set against the respective normalization model (c.f. Equation \ref{eqn:confidence}).}
\label{tab:extrinsic:confusion:matrix}
    \centering
    \scalebox{0.75}
    {
    \begin{tabular}{l|c|c|c|c|c|c|}
    \hline
     & \multicolumn{6}{c|}{\textbf{Normalization Model}}\\ \cline{2-7}
     \textbf{Hate detection method} &  \textbf{FGST} &  \textbf{Bias} & \textbf{ST} & \textbf{DRG} &\textbf{NPTCA} & \textbf{\name-HIR}\\\hline \hline
    \citet{waseem-hovy-2016-hateful} & $0.00$ & $0.03$ & ${0.03}$ &$-0.02$ &$-0.04$ & ${0.03}$ \\
    \citet{davidson2019racial} & $0.04$ & $0.00$ & $0.00$ & $0.35$ & $0.21$ & $0.26$ \\
    \citet{founta2018unified} & $0.04$ & $-0.01$ &$0.07$ &${0.23}$ &$0.04$ &$0.03$ \\ \hline
    \end{tabular}
    }
\vspace{-9pt}
\end{table}
Table \ref{tab:extrinsic:confusion:matrix} reports the average difference in confidence scores for \name-{HIR} and other baselines\footnote{Since we can't generate the dictionary-driven normalized samples for test set in $S_B$ in the absence of hateful spans (c.f. Section \ref{sec:baseline}), we do not include it in our extrinsic evaluation.}. We observe the consistency of \name-{HIR} over other normalization models as it obtains confidence scores $>0$ across all three hate detection methods. In comparison, Style Transfer DRG gives the highest performance in two cases but reports a negative score in the third. {It should be noted that all existing hate speech datasets are highly skewed against hateful labels \cite{founta2018large}. Thus, detection models are prone to underperform for the hate class. Having considered that, extrinsic evaluation aimed at pointing out the general pattern in reduction of hatefulness post-normalization. Such behavior is expected to manifest irrespective of the hate detection model deployed for production.}

\subsection{Error Analysis}
Though \name\ performs better than other baselines across different evaluations, it does commit mistakes. Table \ref{tab:error:strength} lists a few examples along with their original and predicted hate intensity scores. For the first example, \name-{HIP} correctly predicts the hate intensity score as $10$; however, for the other three examples, it wrongly predicts the score. \name-{HIP} assigns lesser intensity score to the second utterance -- the hatred is conveyed by a single word `\textit{qu**r},' and most likely, our model does not assign adequate focus on it in the presence of the positive phrase `happy birthday'. In the third example, the predicted high score is possibly because of the presence of three words -`\textit{bi*ch}', `\textit{damn}', and `\textit{unfollow}'. In comparison, we hypothesize that our model learns to associate color as a cue for hate, thus predicting a high hate intensity score for the last example. 
\begin{table*}
    \centering
    \caption{A few erroneous examples of \name-HIR-generated vs. gold normalized texts.}
    \label{tab:error:normalization}
    
    {
    \begin{tabular}{l|c|p{45em}}
    \hline
  \bf \# & \bf{Type}  & \hfil\bf{Example} \\ \hline \hline
   \multirow{3}{*}{1.} & Original & \em \#LateNightThoughts how many Congressman \{d***s did women s**k\} to finally gain voting rights \\ \cdashline{2-3}
   & Reference & \em \#LateNightThoughts how many Congressman \{did women approach\} to finally gain voting rights \\ \cdashline{2-3}
   & Generated & \em \#LateNightThoughts  how many Congressman \{did women s**k\} to finally gain voting rights \\ \hline
   
   \multirow{3}{*}{2.} & Original & \em \{S**s are half breed trash\}. No \{filthy native\} should be allowed to speak to any European. \\ \cdashline{2-3}
   & Reference & \em No \{native\} should be allowed to speak to any European. \\ \cdashline{2-3}
   & Generated & \em \{Mexicans are t**h\}. No \{disgusting native\} should be allowed to speak to any person. \\ \hline
\end{tabular}
}
\vspace{-5mm}
\end{table*} 
\vspace{-4mm}
    
\begin{table}[!h]
    \centering
     \caption{Gold and predicted values for hate intensity prediction using our proposed model. }
    \label{tab:error:strength}
{
    \begin{tabular}{c|p{16em}|c|c}
    \hline
     \# & \multicolumn{1}{c|}{\bf Example} & \bf Gold & \bf Pred \\ \hline \hline
     1. & \em b***h you're a grown a** women @user & 10 & 10 \\ \hline
     2. & \em @user happy birthday qu**r & 8 & 4 \\ \hline
     3. & \em @user @user @user Man b***h am not gonna do a damn thing , if u don't like what I got to say then unfollow me  & 4 & 10 \\ \hline
     4. & \em firefighter is a job for white people & 2 & 8 \\ \hline
\end{tabular}
}
 \vspace{-5mm}
\end{table}
We also encounter some misclassifications in hateful span identification, as shown in Table \ref{tab:error:span}. In span identification, \name-{HIP} commits three kinds of mistakes -- (a) removes a token from hateful span during prediction; (b) includes additional tokens as part of the hateful span; and (c) classifies non-hate spans as hate. The first kind of mistake can be observed from examples 1 and 2 in Table \ref{tab:error:span}, while the second type is shown in example 3. Similarly,  example 4 reflects the third type of mistake.

Finally, Table \ref{tab:error:normalization}  lists a couple of examples considering the hate normalization task. In the first example, \name-{HIR} partially normalizes the hate sample, and as a consequence, the generated text has low adequacy and fluency scores. Similarly, in the second, we see: (a) the generated sentence is relatively more fluent than the first; and (b) the intensity value is on the higher side due to the presence of phrases `\textit{Mexicans are t**h}' and `\textit{disgusting}.' We argue that the problem of partial normalization can be effectively addressed with more volume and variety of training samples.

\begin{table*}[t!]
    \centering
        \caption{The left part of the table shows an example of human evaluation where we give the generated text to human annotators to evaluate the output of each competing model on three parameters -- intensity, fluency, and adequacy.The results are an average from all the responses received for this example. The right part of the table shows the average scores of all three parameters  across all responses (i.e overall system averages). Intensity ranges from $1-10$, Fluency and Adequacy from $1-5$.
    }
    \label{tab:human:evaluation}
    \scalebox{0.95}{
    \begin{tabular}{l|p{25em}|c|c|c c c|c|c}
    \multicolumn{6}{c}{} & \multicolumn{3}{c}{\textbf{Overall Average}} \\ \cline{1-5} \cline{7-9} 
    \multicolumn{1}{c|}{\bf Model}  & \multicolumn{1}{c|}{\bf Tweet} & \bf Int $\downarrow$ & \bf Flu $\uparrow$ & \bf Adq $\uparrow$ & & \bf Int $\downarrow$ & \bf Flu $\uparrow$ & \bf Adq $\uparrow$  \\ \cline{1-5} \cline{7-9} 
    
    \cline{1-5} \cline{7-9}
    Original & \em @user B*** you shouldn't be putting it out there c*** nu**et & 6 & - & - & & 7.7 & - & -\\ \cdashline{1-5} \cdashline{7-9} 
    Dictionary & \em @user b*** you shouldn't be putting it out & 5.72 & 3.45 & 3.45 & & 4.38 & 1.08 & 3.09 \\ \cdashline{1-5} \cdashline{7-9}
    FGST & \em you shouldn't be putting it out there woman nu**et & 4.72 & 3.82 & 3.82 & & 5.36  & 4.08 & 3.61 \\ \cdashline{1-5} \cdashline{7-9}
    Bias Neut & \em @user you shouldn't be putting it & 3.28 & 2.45 & 2.64 & & 6.54 & 3.15 & 3.6\\  \cdashline{1-5} \cdashline{7-9} 
    \name-{HSR} & \em @user you shouldn't be putting it out there & 2.36 & 4.54  & 3.90 & & \bf 4.08  & \bf 4.38 & \bf 4.16 \\ \cline{1-5} \cline{7-9}
    \cline{1-5} \cline{7-9}
    \end{tabular}}
    \vspace{-4mm}
\end{table*}

 \begin{table*}
    \centering
     \caption{Gold and predicted spans for HSI, with spans indicated by blue text. The green text in the second column indicates misprediction -- either new tokens being inserted into the span or gold tokens being excluded from the predicted span.}
    \label{tab:error:span}
    \scalebox{0.85}{
    \begin{tabular}{c|p{26em}|p{26em}}
    \hline
   \# & \hfil\bf{Gold Span} & \hfil\bf{Predicted Span}\\ \hline \hline
    1. & {\em Why do girls follow me ? All I do is make fun of them \{\textcolor{blue}{for how dumb they are}\}\textcolor{red}{$_{Span}$} \{ \textcolor{blue}{\#WomenS**}\}\textcolor{red}{$_{Span}$}} & 
    {\em Why do girls follow me ? All I do is make fun of them \textcolor{green!40!black}{for} \{\textcolor{blue}{how dumb they are }\}\textcolor{red}{$_{Span}$} \{\textcolor{blue}{\#WomenS**}\}\textcolor{red}{$_{Span}$}}  \\ \hline
    2. & {\em \{\textcolor{blue}{go f*** yourself you stupid ugly c***}\}\textcolor{red}{$_{Span}$} @user} &  {\em \textcolor{green!40!black}{go f*** yourself} \{\textcolor{blue}{you stupid ugly c***}\}\textcolor{red}{$_{Span}$} @user}      \\ \hline
    3. & {\em Blac Chyna is \{\textcolor{blue}{straight trash and an abomination to women}\}\textcolor{red}{{$_{Span}$}} everywhere . Dont @ me cause I dont care. \{\textcolor{blue}{H** responsibly , b***es .}\}\textcolor{red}{$_{Span}$}}     &  {\em Blac Chyna is \{\textcolor{blue}{straight trash and an abomination to women}\}\textcolor{red}{{$_{Span}$}} everywhere . \{\textcolor{green!40!black}{Dont @ me cause I dont care .}\}\textcolor{red}{$_{Span}$} \{\textcolor{blue}{H** responsibly , b***es .}\}\textcolor{red}{$_{Span}$}}      \\ \hline
    4. & {\em okay \{\textcolor{blue}{b***h, f**k off}\}\textcolor{red}{$_{Span}$} its not your business \{\textcolor{blue}{fall in a hole and get a**l f***d by satans horn! B***H}\}\textcolor{red}{$_{Span}$} @user} &  {\em \{\textcolor{green!40!black}{okay} \textcolor{blue}{b***h}\}\textcolor{red}{$_{Span}$}, \textcolor{green!40!black}{f**k off} its not your business \{\textcolor{blue}{fall in a hole and get a**l f***d by satans horn! B***H} \textcolor{green!40!black}{@user} \}\textcolor{red}{$_{Span}$}}      \\ \hline
\end{tabular}}
 \vspace{-2mm}
\end{table*}
\subsection{Human Evaluation} \label{sec:human}
In order to check the viability of the overall system, aided by Wipro AI, we perform a human evaluation and assess the generated text from \name-{HIR} and other baselines. We prepare a questionnaire to measure the qualitativeness of the generated texts. Evaluation instructions are listed in Appendix \ref{app:human_eval}.
For a subset of hateful samples from our dataset, the human annotators are provided with four outputs corresponding to four high performing hate normalization systems, i.e., Dictionary-based, Bias Neutralization, FGST, and \name. To reduce bias, we anonymize the systems, randomly shuffled the outputs, and labeled them as A, B, C, D. Given an original sample, we ask $20$ human annotators\footnote{Among $20$ annotators, $10$ were male, and $10$ were female. The age of all the annotators ranged between 25-40 years. All of them were social media savvy.} to evaluate the generated texts on three dimensions -- {\em hate intensity}, {\em adequacy}, and {\em fluency} \cite{pb:mt:book}. Adequacy (the higher, the better) measures the semantic perseverance in the generated text, while fluency (the higher, the better) refers to the linguistic smoothness in the target language. We modify the definition of adequacy to adopt it for the hate speech normalization task. A predicted sentence that does not reduce hate intensity is considered inadequate and should have a lower adequacy score. We also provide the intensity of the original sample for reference. Finally, we aggregate the average intensity, adequacy, and fluency scores, as shown in Table \ref{tab:human:evaluation}. We present the average scores for one sample and the overall average scores across all samples. On average, \name-{HIR} outperforms others, with hate intensity of $4.08$, fluency of $4.38$, and adequacy of $4.16$.

\subsection{Platform-Independence Analysis}
To further check the robustness of our tool, we extend the human evaluation to a cross-platform analysis of \name. We evaluate on randomly selected 100 samples from
Reddit, GAB and Facebook hateful posts, obtained from \cite{qian-etal-2019-benchmark} (for first two) and  \cite{chung-etal-2019-conan}. We employ the same set of annotators and annotation process mentioned in Section \ref{sec:human} to evaluate the quality of \name\ and two best baselines in terms of intensity, adequacy, and fluency. Table \ref{tab:cross:evaluation} shows that \name\ performs convincing well compared to others across platforms.
\begin{table}[!t]
    \centering
    \caption{Human evaluation on three other platforms.}
    \label{tab:cross:evaluation}
    \resizebox{\columnwidth}{!}{
   
    \begin{tabular}{l|c|c|c|c|c|c|c|c|c}
     \hline
         & \multicolumn{3}{c|}{\bf Reddit \cite{qian-etal-2019-benchmark}} & \multicolumn{3}{c|}{\bf GAB \cite{qian-etal-2019-benchmark}} & \multicolumn{3}{c}{\bf Facebook \cite{chung-etal-2019-conan}} \\ \cline{2-10}
         & \bf Int $\downarrow$ & \bf Flu $\uparrow$ & \bf Adq $\uparrow$  & \bf Int $\downarrow$ & \bf Flu $\uparrow$ & \bf Adq $\uparrow$ & \bf Int $\downarrow$ & \bf Flu $\uparrow$ & \bf Adq $\uparrow$ \\ \hline \hline
         \textbf{FGST} & 5.12 & 2.26 & 1.63 & 5.70  & 2.32 & 1.43 & 6.08 & 2.9  & 1.45 \\
         \textbf{Bias}  & 3.28 & 2.02 & 1.01 & 3.89 & 1.82 & 1.02 & 6.47 & 2.41 & 1.06 \\
          \textbf{\name} & 3.25 & 3.8 & 1.92 & 3.29 & 4.25 & 2.71 & 3.2 & 4.05 & 2.6 \\ \hline
    \end{tabular}}
    \vspace{-5mm}
\end{table}
\subsection{Deployment Details}
\label{sec:browser}
As hate normalization aims to be deployed as a prompting system, we develop an interactive web interface for the same. The web service analyzes the composed text on the go; it reports the intensity of hate and upon finding the text hateful ($\phi>\tau$) it suggests a normalized text as an alternative. The web interface is developed in Flask\footnote{\url{https://palletsprojects.com/p/flask/}}, and works in an auto-complete fashion.  
In Figure \ref{fig:demo:snapshots} (Appendix \ref{app:web_tool}), we show the snapshots of the tool for four scenarios -- \textit{no-hate} ($\phi\sim0$), \textit{low-hate}($\phi\le5$), \textit{mild-hate} ($\phi\le7$), and \textit{extreme-hate} ($\phi>7$). The prototype is being rigorously tested for consistency and scalability by the Wipro AI team. Details of in the wild evaluation are listed in Appendix \ref{app:web_tool}. A demo video of our tool is available on our Github.\footnote{{A demo video of our tool \underline{ \url{https://github.com/LCS2-IIITD/Hate_Norm}}}}

\section{Related work}
\label{sec:related work}
\textbf{Hate Speech Detection:} From the simple logistic regression to \cite{waseem-hovy-2016-hateful,davidson2019racial} deep learning-based models \cite{Badjatiya, kiela2020hateful}, the work on hate speech has diversified in volume \cite{cao-lee-2020-hategan, founta2018large}, languages \cite{mubarak-etal-2017-abusive}, granularity and variety\cite{10.1007/978-3-030-75762-5_55}. \citet{10.1145/3232676}, and  \citet{Poletto2020ResourcesAB} put together a detailed survey of various hate speech detection methods and their shortcomings. In this work, we do not propose any new hate speech detection method. Our framework comes into use once a speech is detected as hateful. Thus, hate speech detection underpins the work of hate speech normalization.

\textbf{Rephrasing Hateful Text:}
In the context of rephrasing offensive text, \citet{su-etal-2017-rephrasing} led the initial work by building a rule-based system (29 hand-crafted rules) for rephrasing profane Chinese phrases/terms. We, too, tested such a tf-idf based mapping approach, but the limitations of such rule-based are in terms of out of vocabulary (OOV) phrases. In another work, \citet{santos2018fighting} built an unsupervised text-transfer model for tackling offensive language across social media platforms. In the latest work on unsupervised text transfer for profanity, \citet{tran-etal-2020-towards} employed a vocabulary-based approach to determine if a sentence should be considered as profane and then proceed to generate its non-offensive version. The works by both \citet{tran-etal-2020-towards} and \citet{santos2018fighting} are similar to our use case. However, they differ in that they employed an unsupervised approach that depended on classifiers or lexicons to capture the offensive/profane text and also aim at 180$^{\circ}$ transformation into a non-offensive one. In contrast, our aim is the reduction of hate and not the complete absence of it. On the line of negativity reduction, \citet{madaan2020politeness} employed a tag-based approach for increasing the politeness of a question/answer (query) pair. \citet{neutralizing:news:bias:2020} adopted the similar tag-based approach but for opinionated news sentences, which is closer to our task than a Q\&A-based setup. Starting with \citet{xu2018unpaired}, several studies (both supervised and unsupervised) showed successful rephrasing of a sentence by modifying its sentiment attribute. These methods \cite{dai-etal-2019-style,NIPS2017_2d2c8394, li-etal-2018-delete,reid-zhong-2021-lewis,10.5555/3305381.3305545} largely disentangled the sentiment attribute and then relied on a combination of select rephrase and attention mechanism to generate an output sequence. For our use case, we hope to learn the hateful-span attributes. Our experiments show that alternatively reducing bias or sentiment is inadequate for reducing hate intensity due to the subjective nature of hate speech. Additionally, we observed that unsupervised style transfer methods that rely on sizeable monolithic corpus hardly perform well when trained on low-volume datasets such as ours.

\section{Conclusion and Future Scopes}
To combat the severity of online hate speech, we proposed an alternative solution by introducing the novel task of \textit{hate speech normalization}. To this end, we proposed \name, a neural hate speech normalizer. We collected and manually annotated a new dataset to support the task. We performed exhaustive evaluations to establish the model's efficacy under different settings. 

We observed two major challenges for the hate normalization task -- first, the lack of parallel data to train more sophisticated generative models, and second, the presence of implicit hate in samples. Though the first hurdle can be addressed (albeit expensive) by annotating more samples, handling the implicit hate is cumbersome \cite{elsherief-2021-latent-hatred}. In the current work, we skipped over the implicit hateful samples due to the absence of explicit hate spans. In the future, we would like to put in rigorous effort to handle such cases and increase the size of the dataset. Additionally,  it would be interesting to see how \name\ can be extended to non-English texts.

\begin{acks}
The authors would like to acknowledge the support of the Prime Minister Doctoral Fellowship (SERB India), Ramanujan Fellowship (SERB, India), Infosys Centre for AI (CAI) at IIIT-Delhi, and ihub-Anubhuti-iiitd Foundation set up under the NM-ICPS scheme of the DST,India. We would also like to thank our industry partner Wipro AI. Wipro is an Indian multinational conglomerate with diverse businesses, coordinated the field study for possible deployment. We acknowledge the support of Shivam Sharma, Technical Lead, Wipro AI for the same. We thank all the human subjects for their help in evaluating our tool.
\end{acks}

\bibliographystyle{ACM-Reference-Format}
\bibliography{main}

\newpage
\appendix
\section*{Appendix}
\begin{figure*}[h!]
    \centering
    \subfloat[No-hate]
    {
    \includegraphics[width=0.22\textwidth]{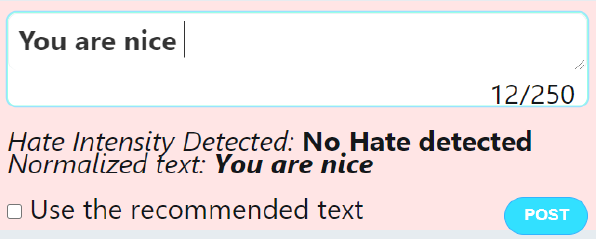}
    } \hspace{0.1em}
    \subfloat[Low-hate]
    {
    \includegraphics[width=0.22\textwidth]{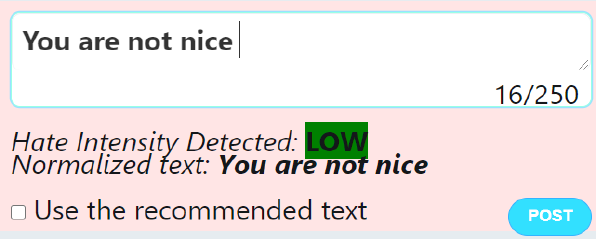}
    } \hspace{0.1em}
    \subfloat[Mild-hate]
    {
    \includegraphics[width=0.22\textwidth]{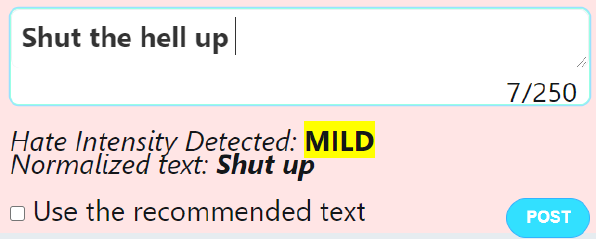}
    }\hspace{0.1em}
    \subfloat[Extreme-hate]
    {
    \includegraphics[width=0.22\textwidth]{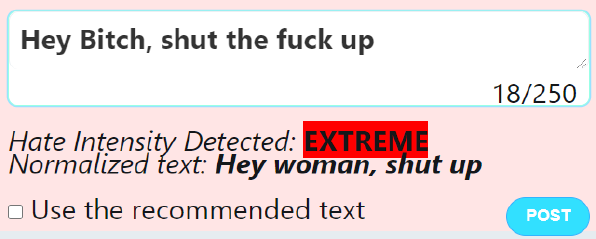}
    }
    \caption{Snapshots of the web extension for four scenarios. \name\ generates normalized text only if $\phi_t > \tau$. The web framework detects hate as the user types in, and if any $\phi_t > \tau.$, it shows the level of hate that is detected in the current text, and then recommends the normalized text to the user.}
    \label{fig:demo:snapshots}
    \vspace{-4mm}
\end{figure*}
\section{Ethical Implications}
\label{app:ethics}
As pointed out by \cite{jin2020deep}, style transfer techniques in NLP can be a force for both good and bad. While, on the one hand, they help turn the Internet into a more positive space \cite{santos2018fighting}, they can also be used to propagate a false/malicious narrative. Reiterating that, our research aims not to support the hateful users but rather to study an alternative approach to prompt users towards a less toxic enunciation of their opinions. Without forcing the users to jump from a negative to a positive space directly, we operate on the space in between as a stepping stone and study the theoretical effectiveness of such a technique (the practical effect is in parts highlighted by human evaluation in Section \ref{sec:human}). 
Keeping in mind the nefarious use of the normalized hate speech (without the knowledge of online platforms about such a normalization), we will not be publicly releasing the parallel corpus. However, it shall be made available to researchers subject to their request for fair usage.

\section{Features for Virality Prediction}
\label{app:viral}
The text-based features employed for predicting comment engagement (aka virality) of the data samples are listed below:
\begin{itemize}[noitemsep, leftmargin=*]
\item Complexity: Uniqueness of terms introduced as measured by the logarithm of term-frequency in the test samples.
\item Readability: LIX and RIX readability scores\footnote{\url{https://readable.com/blog/the-lix-and-rix-readability-formulas/}}. 
\item Informativeness: Summation of the Tf-idf vectors of the words in the sample.   
\item Polarity: Overall SentiNet\footnote{\url{https://pypi.org/project/SentiNet/}} polarity score of the test sample. 
\end{itemize}

\section{Dataset Annotation Guideline}
\label{app:annotation_guideline}
For this experiment, we followed the definition proposed by \cite{waseem-hovy-2016-hateful} for hate speech and marked the hate span if it consists of any of the following explicit mentions:
\begin{itemize}[noitemsep, leftmargin=*]
    \item A sexist or racist slur term, or an abusive term directly attacking a minority group/individual.
    \item A phrase that advocated violent action or hate crime against a group/individual.
    \item Negatively stereotyping a group/individual with unfounded claims or false criminal accusations.
    \item Hashtag(s) supporting one or more of the points as mentioned earlier. 
\end{itemize}
Additionally, the hate intensity of a sample was marked on a scale of $1-10$, $10$ being the highest based on:
\begin{itemize}[noitemsep, leftmargin=*]
\item Score$[8-10]$: The sample promotes hate crime and calls for violence against the individual/group.
\item Score$[6-7]$: The sample is mainly composed of sexist/racist terms or portrays a sense of gender/racial superiority on the part of the person sharing the sample.
\item Score$[4-5]$: Mainly consists of offensive hashtags, or most hateful phrases are in the form of offensive hashtags.
\item Score$[1-3]$: The sample uses dark humor or implicit hateful term.
\end{itemize}

\section{Algorithm}
\label{app:algo}
Algorithm \ref{alg:hsr} shows the learning protocol of \name.
\begin{algorithm}[!th] 
 \small
\caption{Learning \name} 
\label{alg:hsr} 

\textbf{Input:} A set of hateful samples $T=\{t_1,..., t_m\}$; $\forall {j} \quad \phi_{t_j} \ge \tau$ \\
\textbf{Output:} A set of normalized samples $T^\prime = \{t^\prime_1, ..., t^\prime_m\}$; $\forall {j} \quad \phi_{t^\prime_j} < \tau$ \\
\begin{algorithmic} 
    \REQUIRE \name($T, \tau$)
    \STATE Pre-train \texttt{HSI}($T$) and \texttt{HIP}($T$)
    \REPEAT
    \FORALL {$t \in T$}
    \STATE $[\langle s_1, e_1\rangle,..,\langle s_k, e_k\rangle]$ $\gets$ \texttt{HSI}($t$) \hspace{2mm}\textcolor{blue}{\em $\triangleright$ Get indices of hate spans.}
    \STATE $t^\prime \gets$ \texttt{HIR}($t,\langle s_1, e_1\rangle,..,\langle s_k, e_k\rangle$) \hspace{1mm}\textcolor{blue}{\em $\triangleright$ Generator: Get normalized text.}
    \STATE $\phi_{t^\prime} \gets$ \texttt{HIP}($t^\prime$) \textcolor{blue}{\em $\qquad \triangleright$ Discriminator: Get the intensity of the normalized text.}
    \STATE $R_{t^\prime} = \tau - \phi_{t^\prime}$
    \textcolor{blue}{\em $\qquad \quad \triangleright$ Calculate reward for the generator - +ve, if intensity less than threshold, -ve, otherwise.}
    \STATE $\mathcal{L} = \ell + (1 - R)$  \textcolor{blue}{\em $\qquad \triangleright$ $\ell$ = Generator loss}
    \STATE \texttt{HIR} $\gets$ \texttt{Backpropagation}($\mathcal{L}$)
    \ENDFOR
    \UNTIL {<termination-condition>}
    \RETURN $T^\prime$
\end{algorithmic}
\vspace{0.5em}
\begin{algorithmic} 
    \REQUIRE \texttt{HIP}($t$)
    \STATE $H^\prime \gets$ \texttt{BiLSTM}(\texttt{BiLSTM}($t$))
    \STATE $\alpha \gets$ \texttt{Attention}($H^\prime$)
    \STATE $H = H^\prime \cdot \alpha$
    \STATE $\phi_t \gets$ \texttt{Linear}($H$)
    \RETURN $\phi_t$ \textcolor{blue}{\em $\qquad \qquad \qquad \quad \triangleright$ Hate intensity of the tweet.}
\end{algorithmic}
\vspace{0.5em}
\begin{algorithmic} 
    \REQUIRE \texttt{HSI}($t$)
    \STATE $\langle w_1, ..., w_n\rangle = t$
    \STATE $h_1, ..., h_n  \gets$ \texttt{BiLSTM}($\langle w_1, ..., w_n\rangle$)
    \STATE $p_1, ..., p_n \gets$ \texttt{CRF}($h_1, .., h_n$) \textcolor{blue}{$\qquad \triangleright \forall_{i} \quad p_i \in \{B, I, O\}$}
    \STATE $[\langle s_1, e_1\rangle,..,\langle s_k, e_k\rangle] \gets$ GetSpans($p_1,..,p_n$) \textcolor{blue}{\em $\quad \triangleright$ <$s, e$> are the start and end indices of the hate span.}
    \RETURN $[\langle s_1, e_1\rangle,..,\langle s_k, e_k\rangle]$
\end{algorithmic}
\vspace{0.5em}
\begin{algorithmic} 
    \REQUIRE \tt{HIR}($t,\langle s_1, e_1\rangle,..,\langle s_k, e_k\rangle$)
    \STATE $t^\prime \gets t$
    \FORALL{$\langle s_i, e_i\rangle$}
        \STATE \textcolor{blue}{\em $\triangleright$ For each identified hate span.}
        \STATE $H \gets$ \texttt{BART-encoder}($t$)
        \STATE $t^\prime_{[s_i:e_i]} \gets$ \texttt{BART-decoder}($H, \langle s_i, e_i\rangle$) \textcolor{blue}{\em $\quad \triangleright$ Update hate span with normalized text.}
    \ENDFOR
    \RETURN $t^\prime$
\end{algorithmic}
\end{algorithm}
 \setlength{\textfloatsep}{1pt}

\section{Human Evaluation Guideline}
\label{app:human_eval}
After reading the original sample and the normalized counterparts, the annotators provided their input in the form of the following:
\begin{itemize}[noitemsep, leftmargin=*]
    \item Intensity: The annotators assigned a hateful intensity score to each generated output on a scale of [1,10], $10$ being the highest intensity. 
    \item Fluency: To understand how well constructed and readable the generated text is, the annotators scored each generated text on a range of [1,5], $5$ being the highest fluency.
    \item Adequacy: Additionally, to provide an idea of whether the desired meaning can be interpreted from the output, the annotators were asked to score each generated text on a range of [1,5], $5$ being the highest. Since our task aims to perform normalization and not the conversion of hate to non-hate, if a sentence changes the sample's polarity, then that can also be taken as a negative case from our intended perspective. The annotators were informed before that a normalized sample with its polarity reversed would have a minimum ($1$) adequacy even if it is fluent.
\end{itemize}
For reference, the intensity of the original sample was provided. The fluency and adequacy of the original sample were considered highest. To reduce the annotator's bias, the annotators were unaware of which output text represented our system. 

\section{Hyperparameters}
The models make use of Tensorflow 2.0 and Transformer 4.5.1 with Python 3 libraries, trained on Google Colab with Tesla P100-PCIE-16GB GPU.
\label{app:hyper}
\begin{itemize}[noitemsep, leftmargin=*]
    \item For the HIP model, we employ a single Bi-LSTM layer (\textit{hdim}=$512$) followed by a layer of self-attention. For this model, use the MSE loss and the Adam optimizer with linear activation. The input embedding is BERT \textit{dim}=$768$. The model is trained for $10$ epochs with a batch size of $32$.
    \item For the HSI, we employ a $2$ layer Bi-LSTM (\textit{hdim}=$512$), using a batch size of $32$, and the RELU activation function. As the final layer for the Span Model is a CRF, we take crf\_loss and crf\_accuracy as our loss and accuracy metric respectively, and Adam optimizer. The model is trained for $5$ epochs with a batch size of $32$.
    \item For the HIR, we employ Facebook's BART-base module, with a cross-entropy loss, RELU activation and Adam optimizer.
\end{itemize}

\section{Deployment Details}
\label{app:web_tool}
The proposed web-interface can easily be made to work in cross platform settings to proactively curb hate speech, and the screenshot for the same is provided in Figure \ref{fig:demo:snapshots}.

Again we evaluate the tool in the wild by asking $25$ participants to write random hateful content on their own and rate the tool's output. We extend the tool by including the interface of the Bias Neutralization baseline -- the tool shows the outputs of \name\ and the baseline for a given input. Each participant is asked to input hateful content, assign an original intensity score, and subsequently evaluate the tools' outputs considering fluency, adequacy, and intensity. The normalization methods are anonymized for the participants. In total, we obtain $100$ input samples, with an average normalized intensity score of $3.24$. Similar to earlier observations, we notice that \name\ results in more fluent sentences with higher reduction in intensity. This in-the-wild evaluation further supplements that \name\ is not restricted to our dataset. 

\section{Extrinsic Evaluation}
\label{app:ext_eval}
The distribution of hateful and non-hateful samples for the three hate speech detection models we use in our extrinsic evaluation.
\begin{table}[h!]
    \centering
    \caption{The hate vs non-hate tweet ratio for the hate classification. The data is collected from the respective methods.}
    \label{tab:hate_detection_stats}
    {
    \begin{tabular}{l|c|c}
    \hline
          \multicolumn{1}{c|}{\textbf{Model}} & \textbf{\# of tweets} & \textbf{Hate:Non-Hate}\\ \hline
          \citet{waseem-hovy-2016-hateful} & 12772  & 4750:8022\\
          \citet{davidson2019racial} & 24783 & 20620:4183\\
          \citet{founta2018unified} & 59189 & 13551:45638\\ \hline
    \end{tabular}
}
\end{table}

\end{document}